\documentclass[journal]{IEEEtranTIE}
\usepackage{graphicx}
\usepackage{cite}
\usepackage{picinpar}
\usepackage{amsmath}
\usepackage{url}
\usepackage{flushend}
\usepackage[latin1]{inputenc}
\usepackage{colortbl}
\usepackage{soul}
\usepackage{multirow}
\usepackage{pifont}
\usepackage{color}
\usepackage{alltt}
\usepackage[hidelinks]{hyperref}
\usepackage{enumerate}
\usepackage{siunitx}
\usepackage{breakurl}
\usepackage{pbox}
\usepackage[T1]{fontenc}
\usepackage{cite}

\usepackage{amsmath}
\usepackage{times} 
\usepackage{amssymb}  
\usepackage{subfigure}
\usepackage{bm}
\usepackage{physics}
\usepackage{threeparttable}
\usepackage{gensymb}
\usepackage{fancyhdr}
\usepackage{booktabs}
\usepackage{float}
\usepackage{bigstrut}
\usepackage{multirow}
\usepackage{balance}
\usepackage{url}
\usepackage{bm}
\usepackage{booktabs}
\usepackage{float}
\usepackage{bigstrut}
\usepackage{caption}
\usepackage{amsopn}
\usepackage{epstopdf} 
\usepackage[vlined,ruled]{algorithm2e}
\usepackage[table]{xcolor}
\usepackage[mathlines,switch]{lineno}
\usepackage{placeins}
\usepackage{amsbsy}
\usepackage{amsthm}
\usepackage{amsfonts}
\usepackage{multirow}

\usepackage{booktabs,makecell,multirow}
\usepackage{url}

\begin{document}
\title{A Novel Feature Learning-based Bio-inspired Neural Network for Real-time Collision-free Rescue of Multi-Robot Systems}

\author{
	\vskip 1em
	
	 Junfei Li, \emph{Member,~IEEE},
       Simon X. Yang, \emph{Senior Member,~IEEE},

	\thanks{

		This work was supported in part by the Natural Sciences and Engineering Research Council of Canada. (Corresponding author:Simon X. Yang.)
		
		Junfei Li and Simon  X.  Yang  are  with  the  Advanced  Robotics  and  Intelligent  Systems  Laboratory, School  of Engineering,  University  of Guelph,  Guelph,  ON N1G2W1, Canada. (e-mail: \{jli64;syang\}@uoguelph.ca). 
	}
}

\maketitle

\begin{abstract}
Natural disasters and urban accidents drive the demand for rescue robots to provide safer, faster, and more efficient rescue trajectories. In this paper, a  feature learning-based bio-inspired neural network (FLBBINN) is proposed to quickly generate a heuristic rescue path in complex and dynamic environments, as traditional approaches usually cannot provide a satisfactory solution to real-time responses to sudden environmental changes. The neurodynamic model is incorporated into the feature learning method that can use environmental information to improve path planning strategies. Task assignment and collision-free rescue trajectory are generated through robot poses and the dynamic landscape of neural activity.  A dual-channel scale filter, a neural activity channel, and a secondary distance fusion are employed to extract and filter feature neurons. After completion of the feature learning process, a neurodynamics-based feature matrix is established to quickly generate the new heuristic rescue paths with parameter-driven topological adaptability. The proposed FLBBINN aims to reduce the computational complexity of the neural network-based approach and enable the feature learning method to achieve real-time responses to environmental changes. Several simulations and experiments have been conducted to evaluate the performance of the proposed FLBBINN. The results show that the proposed FLBBINN would significantly improve the speed, efficiency, and optimality for rescue operations.
\end{abstract}

\markboth{This article has been accepted for the IEEE Transactions on Industrial Electronics.  DOI: 10.1109/TIE.2024.3370939}%
{}

\definecolor{limegreen}{rgb}{0.2, 0.8, 0.2}
\definecolor{forestgreen}{rgb}{0.13, 0.55, 0.13}
\definecolor{greenhtml}{rgb}{0.0, 0.5, 0.0}

\section{Introduction}

The application of robotics in public safety and rescue operations has garnered significant attention due to its potential to save lives and mitigate risks \cite{geng2018good}. For instance, unmanned ground vehicles (UGVs) have been widely used for search and rescue operations in urban and wilderness settings \cite{mohamed2019person}. In addition, UGVs can navigate through debris, confined spaces, and rough terrain, facilitating victim detection and information gathering. 
Unmanned aerial vehicles (UAVs) or drones have emerged as valuable tools for search and rescue operations \cite{wu2021reinforcement}. Their ability to cover large areas and access hard-to-reach locations quickly makes them particularly useful for locating missing persons. 
Moreover, marine robots, including unmanned surface vehicles (USVs) and autonomous underwater vehicles (AUVs), have been developed for search and rescue missions at sea \cite{yang2020maritime}. These robots are able to locate and track shipwreck victims, assess oil spills, and monitor underwater infrastructure. 

The concept of rescue robots has gained significant attention due to their potential to save lives in hazardous environments. 
Harikumar \textit{et al.}  \cite{harikumar2018multi} employed multiple UAVs in firefighting operations, as they can endure high temperatures and hazardous environments that pose significant risks to human firefighters.
Nishikawa \textit{et al.}  \cite{cardenas2019design} developed a disaster response robot that could play a crucial role in debris removal, rescue of trapped victims, and support of reconstruction efforts. 
 Sun \textit{et al.} \cite{sun2021bit} developed a dual-arm mobile robot for narrow space and complex rescue operations. Niroui \textit{et al.} \cite{niroui2019deep} proposed a deep reinforcement learning-based rescue robot in unknown urban environments.
In addition, robotic systems have been used to minimize the risk to human operators in various hazardous environment operations\cite{delmerico2019current}. Lin \textit{et al.} \cite{lin2020task} proposed a task framework for rescue robots in a chemical disaster environment, where the potential threat is suddenness, danger, and variability.

The bio-inspired neural network (BINN) is well suited for path planning 
in dynamic environments because BINN can generate the path without prior knowledge of the dynamic environment and any learning procedures. However, BINN is computationally expensive in real-world rescue operations because its computational complexity depends linearly on the size of the neural network. Recently, there has been a trend to use environmental feature learning to improve the performance of path planning \cite{jung2018development,yuan2021knowledge,yuan2022gaussian}. Inspired by feature learning, the rescue process is also capable of providing useful environmental information to improve rescue planning strategies.
 
In this paper, a novel intelligent multi-robot framework is proposed for executing rescue operations in complex and dynamic environments.
A feature learning-based bio-inspired neural network (FLBBINN) is proposed to take advantage of environmental information collected during robotic rescue to extract feature neurons to improve the effectiveness and efficiency of rescue performances.
The real-time assignment of tasks and the collision-free rescue trajectory are generated through the dynamic activity landscape of the neural network.
The term ``real-time" is defined as the robot motion planner responding immediately to the dynamic environment.
To decrease the computational complexity of the neural network-based approach, a neurodynamics-based feature learning technique is proposed to extract feature neurons during rescue process. A dual-channel scale filter, a secondary distance fusion, and a neural activity channel are used to further extract  feature neurons for pruning the neural network. Finally, upon extracting the feature neurons, a topological neurodynamics-based feature matrix is built for facilitating environmental changes and the generation of rapid rescue paths.
The main contributions of this paper are summarized as follows:
\begin{itemize}
		\item  A novel FLBBINN is proposed for executing rescue operations in complex and dynamic environments.
  
            \item A novel feature extraction method incorporated with the dynamic landscape of neural activity is proposed to reduce computational complexity, generate shorter paths, and deal with the slow propagation problem of BINN.

        \item A novel fast heuristic path planning with parameter-driven topological adaptability is proposed based on the neurodynamics-based feature matrix.
	     

\end{itemize}

This paper is organized in the following manner. Section \ref{Realted} provides the related work on the neurodynamics model and conventional BINN.  Section \ref{sec:Problem} gives the problem statement. Section \ref{sec:approach} describes the proposed approaches. Section \ref{sec:simulation} shows the simulation and  comparison results. Section \ref{sec:experiments} provides a real robot experiment.  In Section \ref{sec:conclusion}, the result is briefly summarized.

\section{Related Work}
\label{Realted}
In 1952, Hodgkin and Huxley \cite{hodgkin1952quantitative} introduced an electrical circuit model to represent the membrane potential within a neuronal system. 
Grossberg \cite{grossberg1988nonlinear} formulated a shunting neurodynamics model based on the dynamics of the membrane. The shunting equation can be expressed as follows 
\begin{equation}
\frac{d \zeta_k}{d t}=-A \zeta_k+\left(B-\zeta_k\right) S_{k}^{e}-\left(D+\zeta_k\right) S_{k}^{i},
\label{eq:shunnting}
\end{equation}
where $\zeta_k$ denotes the neural activity of the $k$-th neuron; $S^e_k$ and $S^i_k$ represent the excitatory and inhibitory inputs to the neuron, respectively. The parameters of the shunting equation has been discussed in previous works \cite{Li2021bio}. Parameter $A$ is the passive decay rate and the transient response of the external inputs is exclusively determined by the parameter $A$. Parameters $B$ and $D$ are the upper and lower bounds of neural activity, respectively. Under different input conditions, the neural activity $\zeta_k$  will remain in the region $[-D, B]$. In the BINN, the dynamics of the $k$-th neuron in the neural network is characterized by a shunting equation derived from (\ref{eq:shunnting}). The excitatory input $S_{k}^{e}$ results from the target and the lateral connections from its neighboring neurons, while the inhibitory input $S_{k}^{i}$ results from the obstacles only. Therefore, the neural activity of the $k$-th neuron is denoted as follows
\begin{equation}
\begin{aligned}
\frac{d \zeta_k}{d t}=-A\zeta_k&+(B-\zeta_k)\Bigg([I_k]^++\sum^n_{l=1}w_{kl}[\zeta_l]^+\Bigg)\\
&-(D+\zeta_k)[I_k]^- ,
\end{aligned}
\label{eq:binnE}
\end{equation}
where $\zeta_l$ denotes the neural activity of neurons neighboring the $k$-th neuron; $n$ signifies the number of these neighboring neurons; $[a]^+$ is defined by $[a]^+=\max\left\{a,0\right\}$; and $[a]^-$ is defined as $[a]^-=\max\left\{-a,0\right\}$.  The connection weight $w_{kl}$ is defined as 
\begin{equation}
w_{kl}=f(|kl|)=\left\{
\begin{aligned}
&\mu/|kl|,  \quad  0<|kl|\leq r_0 ,\\ 
&0, \quad \qquad |kl|>r_0,
\end{aligned}
\right.
\label{eq:connection}
\end{equation}
where $|kl|$ denotes the Euclidean distance between the $k$-th and $l$-th neurons and $\mu$ is a positive constant, which is solely determines the connection weight.  The proposed neural network characterized by (\ref{eq:binnE}) guarantees that the positive neural activity can propagate to the whole state space through lateral neural connections, while the negative activity stays locally only, since there is no inhibitory connections among neurons. Consequently, the target displays the maximal and positive neural activity that is capable of propagating globally throughout the neural network. On the contrary, obstacles exert exclusively local effects without any propagation. The BINN has been widely used in various autonomous robots \cite{Li2021bio}. Many complex tasks based on BINN have been studied, such as avoiding collisions with the mobile robot \cite{yang2003real}, cooperative hunting\cite{ni2011bioinspired}, complete coverage path planning \cite{luo2016neural} and target search of multiple AUVs \cite{zhu2014path}.
\begin{algorithm}[t]
\footnotesize
	\LinesNumbered
	\caption{Computing Neural Activity for Path Planning}
	\label{alg:BINN}
	
	\KwIn{Environment information and size $N_x \times N_y$, positions of target $\mathbf{T}_c$}
	\KwOut{dynamic neural activity of neural network $\zeta$}

 \While {rescue tasks not finised}
    {

   	       \For{$x_i=1$ to $N_x$}
	             {
                    \For{$y_i=1$ to $N_y$}
	             {
                       Set targets as the excitatory input $S^e_k$\\
                       Set obstacles as the inhibitory input $S^i_k$\\
                        
                         \For{$x-r_0$ to $x+r_0$}
	             {
                         \For{$y-r_0$ to $y+r_0$}
	                  {
                           \If{ $ \zeta(x,y)	\geq 0$}
			            {
				            $S^e_k= S^e_k +w_{kl}*\zeta(x,y)$\\
	                      }
                            Compute neural activity $\zeta$ by (2)\\ 
                        }
                  }

	           }
	           }
              
    }
  \Return $\zeta$
\end{algorithm}  

However, as shown in Algorithm \ref{alg:BINN}, there are some limitations of traditional BINN in the application of rescue tasks. Firstly, the computational complexity exhibits a linear dependence on the size of the neural network. Therefore, BINN is not suitable for a large-size environment and is computationally expensive for rescue tasks. Secondly, the neuron connection is established in a fixed manner in eight directions. Therefore, the planned path constitutes the shortest connection between neurons under grid discretization, rather than the actual shortest path. Finally, the neuron only has local lateral connections. Therefore, the propagation time may experience an extension in certain complex rescue situations.



\section{Problem Statement}
\label{sec:Problem}
\begin{figure}[h]
\centering
\includegraphics[width=2in]{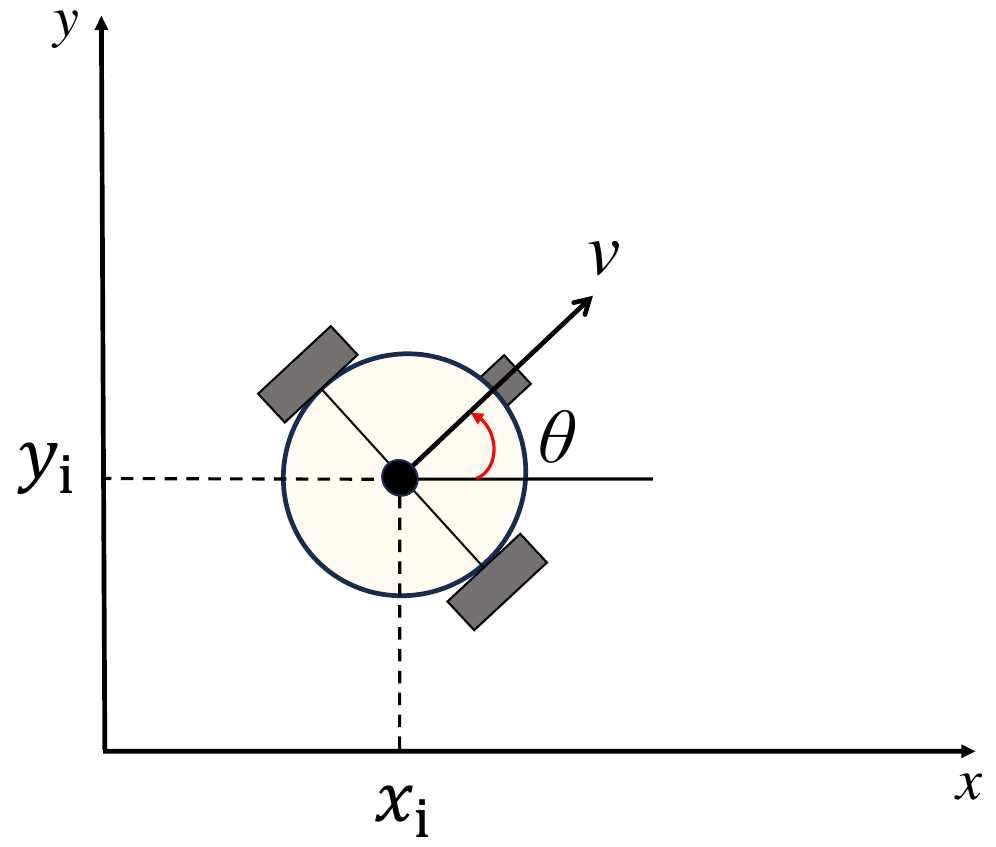}
\caption{Illustration of the robot model.}
\label{fig_robor}
\end{figure}

The time-varying positions of a number of $m$ robots within the 2D environment $W$ at a specific time instant $t$ can be unequivocally ascertained by the spatial coordinates $\mathbf{p}_{i}= (x_i,y_i)$,$i=1,\ldots,m$. Suppose that each robot is considered an omnidirectional robot, which can change direction without delay. As shown in Fig. \ref{fig_robor}, the next location of the $i$-th robot at time instant $t+1$ can be given as 
\begin{equation}
\left(x_{i}\right)_{t+1}=\left(x_{i}\right)_{t}+v_{} \Delta t \cos \left(\theta\right)_{t},
\end{equation}
\begin{equation}
\left(y_{i}\right)_{t+1}=\left(y_{i}\right)_{t}+v_{} \Delta t \sin \left(\theta\right)_{t},
\end{equation}
where $v$ is the speed of the robot; $\theta$ is the moving direction of the robot; and $\Delta t $ is the unit time interval. The robot is equipped with sensors with $360\degree$ visual capability to recognize rescue targets $\mathbf{T}_c$, $c=1,\ldots,q$. 
In addition, there is a sequence of static and moving obstacles in $W$. Let $\mathcal{O}$ be an obstacle scenario. The collision-free area pertaining to static $\mathcal{O}$ can be defined as $\mathcal{O}_{\text {free }}:=\left\{(x,y) \in \mathbb{R}^2: \Gamma >1 \right\}$,
\begin{equation}
\Gamma=\frac{\left(x-x_o\right)^{2}+\left(y-y_o\right)^{2}}{\lambda_o},
\end{equation}
where $(x_o,y_o)$ is the center of the obstacle, and $\lambda_o$ is the size of the obstacle. The regions that meet $\Gamma =1$, $\Gamma >1$, or $\Gamma <1$ denote the surface, exterior, or interior of the obstacle, respectively. If moving obstacle scenarios exist, its time-varying collision-free area is defined as: $\mathcal{O}_{\text {free }}^t:=\left\{(x,y) \in \mathbb{R}^2: \Gamma_t >1\right\},$ where $\Gamma_t$ has the same structure of $\Gamma$. The center of the moving obstacle $(x_o^t,y_o^t)$ is updated with respect to a constant moving speed $v_o$. 

The rescue problem studied in this paper is very different compared to conventional path planning \cite{Li2021bio,yang2003real,ni2011bioinspired,luo2016neural,zhu2014path}. Due to the requirement of high-precision environmental modeling in rescue operations, this necessity for detailed modeling results in a considerable expansion in the size of the neural networks.  In addition, unlike some rescue problems documented in existing literature where the environment is static with obstacles and surroundings that remain constant, the rescue problem in this paper involves dynamic conditions and sudden-change obstacles \cite{harikumar2018multi,cardenas2019design,sun2021bit,niroui2019deep}.  The rescue problem investigated in this paper can be characterized as follows: for a group comprising $m$ robots, considering the initial positions of the robots $\mathbf{p}_i(0)$ with $i=1,\ldots,m$. Since the rescue targets  $\mathbf{T}_c$, $c=1,\ldots,q$ exist in the environment, the $i$-th robot begins to generate a collision-free trajectory, that is, $\mathbf{P}_{\text {Rescue}} \in \mathcal{O}_{\text {free }}$, until it achieves the position of $\mathbf{T}_c$. 

\section{Proposed Approaches}
\label{sec:approach}
\begin{figure*}[t]
\centering
\includegraphics[width=6in]{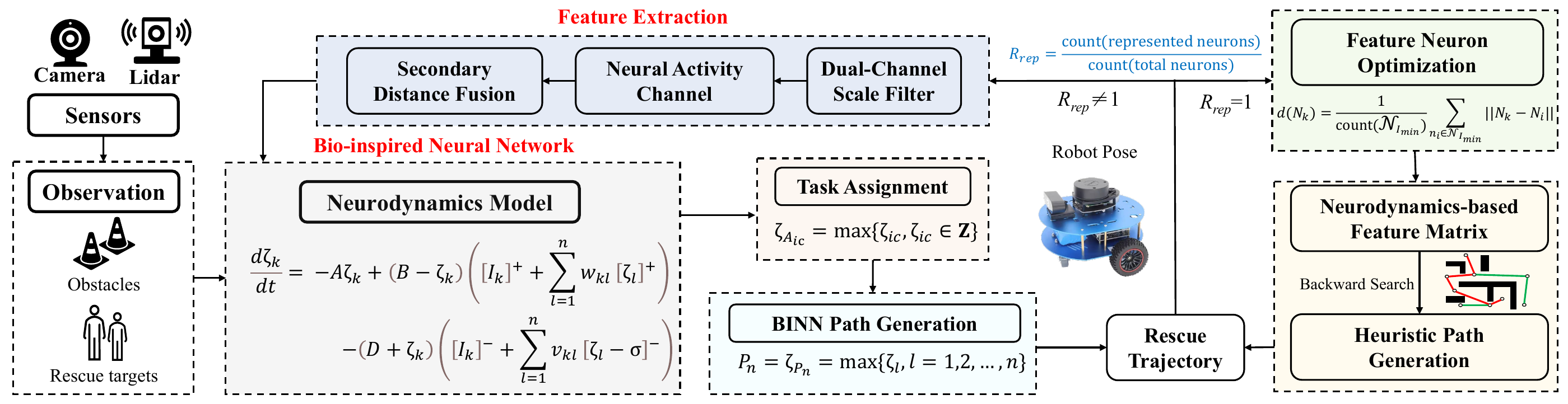}
\caption{Illustrations of the proposed intelligent multi-robot rescue framework.}
\label{fig_framework}
\end{figure*}

In this study, a novel intelligent multi-robot rescue framework is proposed to generate the collision-free rescue paths of multi-robot systems in dynamic and complex environments. By combining the strengths of BINN and the feature learning method, the proposed  FLBBINN aims to reduce computation complexity and enable multi-robot systems to achieve real-time responses to dynamic and complex environments. In the following paragraphs, the key components of the proposed approaches are introduced, as illustrated in Fig. \ref{fig_framework}.

\subsection{Safety and Intelligent Rescue Path Generation}
\begin{figure}[htbp]
\centering
\includegraphics[width=3.4in]{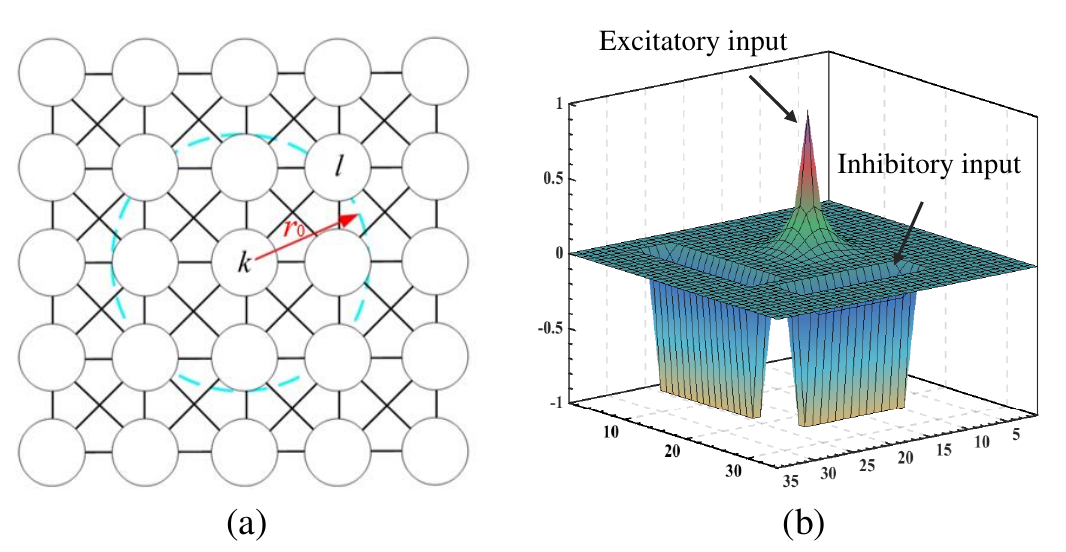}
\caption{ Examples of the bio-inspired neural network. (a) structure of the neural network featuring exclusively local connections; (b) the dynamic landscape of neural activity.}
\label{fig_BINN}
\end{figure}

The proposed neural network structure is shown in Fig. \ref{fig_BINN}(a). Each neuron is one-to-one representing an environmental location. The neuron receptive field is represented by a circle with a radius of $r_0$ and has lateral connections only to its eight neighboring neurons. 
The dynamics of neural activity to the $k$-th neuron within the neural network is delineated by a shunting equation (\ref{eq:binnE}). Due to the complexity and uncertainty of the rescue environment, a safety consideration approach is used to generate the real-time path. The neural activity of $k$-th neuron with safety consideration is denoted as follows
\begin{equation}
\begin{aligned}
\frac{d \zeta_k}{d t}=-A\zeta_k&+(B-\zeta_k)\Bigg([I_k]^++\sum^n_{l=1}w_{kl}[\zeta_l]^+\Bigg)\\
&-(D+\zeta_k)\Bigg([I_k]^-+\sum^n_{l=1}v_{kl}[\zeta_l-\sigma]^-\Bigg),
\end{aligned}
\label{eq:binnE_safe}
\end{equation}
where $\sigma$ represents the threshold for inhibitory lateral neural connections and $v_{kl}$  is the connection weight which can be defined as $v_{kl}=\beta w_{kl}$, where $\beta$ is a positive constant, $\beta \in [0, 1]$.
Compared to (\ref{eq:binnE}), the inhibitory signal $S^i_k$ in (\ref{eq:binnE_safe}) induces negative neural activity, and the term $\sum^n_{l=1}v_{kl}[\zeta_l-\sigma]^-$ facilitates the confinement of negative activity within a limited local region due to the presence of threshold $\sigma$.
Therefore, the excitatory signal $S^e_k$ globally influences the entire state space, while the inhibitory signal $S^i_k$ only has a local effect in a small region, as shown in Fig. \ref{fig_BINN}(b).
The external input $I_{k}$ is defined as
\begin{equation}
I_{k}=\left\{
\begin{aligned}
&E,  \qquad \, \text{if it is a target},\\
&-E,  \quad \text{if it is an obstacle},\\
&0,   \qquad \ \, \text{otherwise},
\end{aligned}
\right.
\end{equation}
where $E$ represents a positive constant. In the event that the corresponding position represents a target, the external input adopts a markedly positive value. However, if the corresponding position constitutes an obstacle, the external input assumes a considerably negative value.
\subsubsection{\textbf{Task Assignment}}
In the beginning, obstacles exhibiting maximum external inhibitory inputs exert influence only within a limited local region, whereas targets with maximum external excitatory inputs have a global impact. The activity values of all robots to each target can be represented by a matrix $\mathbf{Z}$ \cite{zhu2020novel}
\begin{equation}
\begin{array}{c}
\mathbf{Z}=\left[\begin{array}{cccc}
\zeta_{11}  \ldots & \zeta_{a b}
\end{array}\right], \\
\end{array}
\end{equation}
\begin{equation}
\begin{array}{c}
 \zeta_{A_{ ic }}=\max \left\{\zeta_{i c}, \zeta_{i c} \in \mathbf{Z} \right\},
\end{array}
\end{equation}
where $a$ is the number of robots, and $b$ is the number of targets; $\zeta_{ic}$is expressed as the activity value of the $i$-th robot to the $c$-th target; $\zeta_{A_{ ic }}$ indicates the $i$-th robot has the highest activity value  to the $c$-th target compared with other robots.
The task assignment of target $\mathbf{T}_c$ can be defined as
\begin{equation}
\mathbf{T}_c \Rightarrow \left\{ 
\begin{aligned}
 &i\text{-th robot}, \qquad \, \, \text{if $\zeta_{A_{ ic }} > \zeta_{A_{ mc }}$},  \\
&m\text{-th robot},  \quad \,\,\,\,\, \text{if $\zeta_{A_{ ic }} < \zeta_{A_{ mc }}$},  \\
& \text{not assigned} ,  \quad \, \text{if $\zeta_{A_{ ic }} = \zeta_{A_{ mc }}$},   \\
\end{aligned}
\right.
	\label{taskassign}
\end{equation}
where $\zeta_{A_{ ic }}$ is the neural activity of $i$-th robot to $\mathbf{T}_c$ and $\zeta_{A_{ mc }}$ is the neural activity of $m$-th robot to $\mathbf{T}_c$, respectively. If $\zeta_{A_{ ic }} > \zeta_{A_{ mc }}$, the target $\mathbf{T}_c$ is assigned to  $i$-th robot. If $\zeta_{A_{ ic }} < \zeta_{A_{ mc }}$, the target $\mathbf{T}_c$ is assigned to  $m$-th robot. 
If $\zeta_{A_{ ic }} = \zeta_{A_{ mc }}$ then the target is temporarily not assigned to robots.
The assignment of robots to individual targets is determined by comparing the activity values of other robots within their respective neural networks. This approach addresses the issue of redundant robot allocation arising from the sequence of target input.

\subsubsection{\textbf{Path Planning}}
\begin{figure}[htbp]
\centering
\includegraphics[width=1.8in]{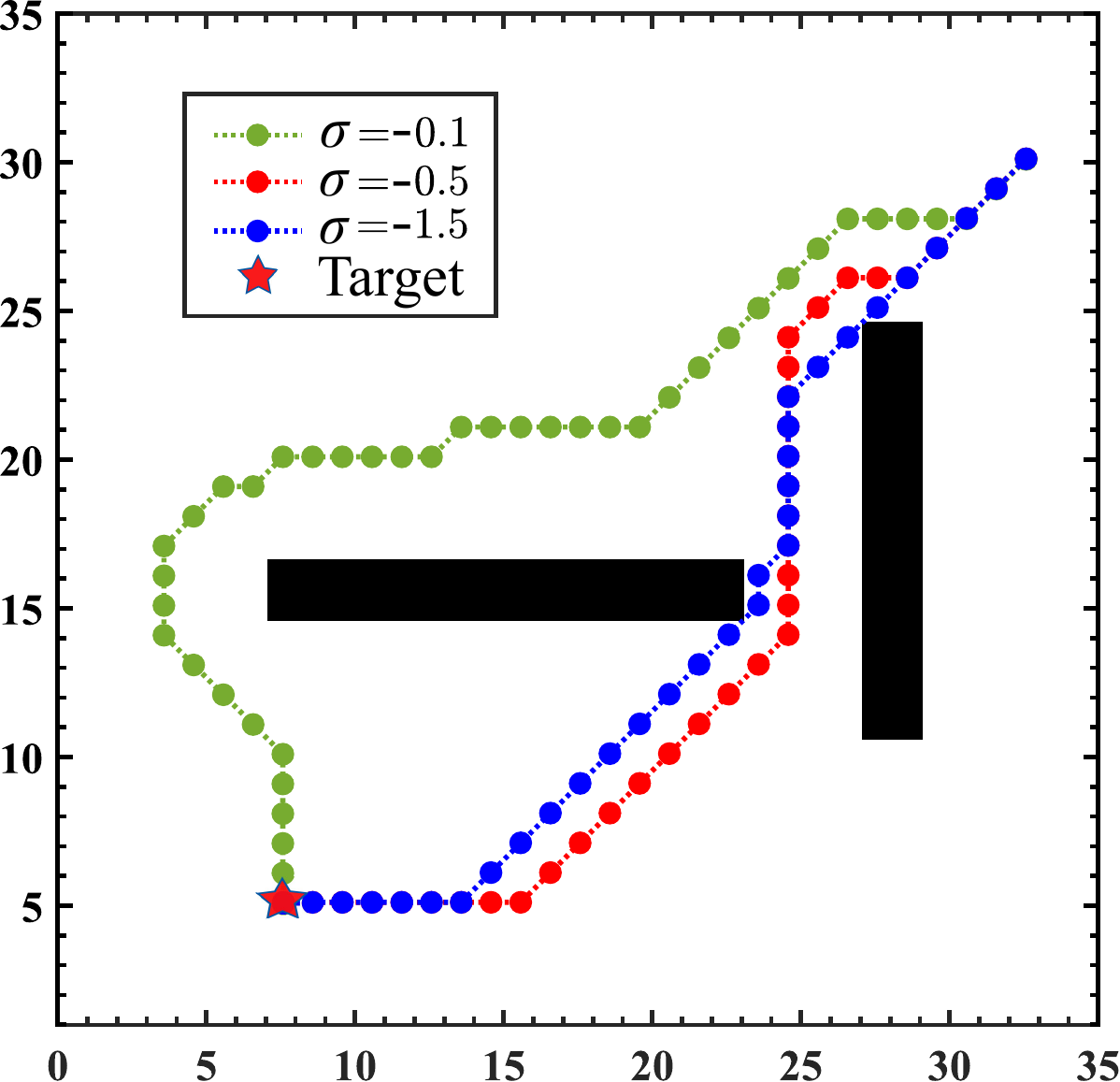}
\caption{Three typical examples of one robot by choosing different parameters $\sigma$ values.}
\label{fig_sigma}
\end{figure}

The command neuron of the next position can be given as
\begin{equation}
\begin{aligned}
P_{n} \Leftarrow \zeta_{P_{n}}=\rm{max}&\left\{ \zeta_l,l=1,2,...,n\right\},&
\label{search_max}
\end{aligned}
\end{equation}
where $P_{n}$ represents the command neuron of the next position in the neural network; $\zeta_{P_{n}}$ represents the neural activity of the command neuron. From (\ref{search_max}), the robot continues to search for the maximum neural activity in its neighborhoods. 
Fig. \ref{fig_sigma} shows three typical examples of a robot by choosing different parameters $\sigma$. 
The blue line shows the motion of the robot when choosing $ \sigma= -1.5$. There is no clearance from the obstacles. It is obvious that the robot clips the corners of the obstacles and runs down the edges of the obstacles. Therefore, possible interactions between robots might cause a collision with obstacles when the number of robots increases. The green line shows a strong obstacle clearance by choosing $\sigma = -0.1$. It shows a very strong clearance from obstacles, which pushes the robot very far from the obstacle. When considering the gap between obstacles is not able to accommodate robots, choosing a small $\sigma$ value is a sub-optimal option, even if it is not the shortest rescue path.
The red line illustrates the movement of robots with moderate obstacle clearance, achieved by selecting $\sigma = -0.5$. The generated trajectory is considered comfortable because it avoids intersecting the corners of obstacles and does not closely follow the contours of the obstacles.

\subsection{ Environmental Feature Learning via Neurodynamics}
The pseudocode of the proposed approach is shown in Algorithm \ref{alg:NN}. In the beginning, the position and direction of the robot moving to the neuron can be obtained as $\mathbf{p} = (x, y, \theta)$. 

\begin{algorithm}[t]
\footnotesize
	\LinesNumbered
	\caption{Extraction and Filtering of Feature Neurons}
	\label{alg:NN}
	
	\KwIn{BINN and pose of robot $\mathbf{p} = (x, y, \theta)$}
	\KwOut{Feature neurons set $N_{F}$ and neurodynamics-based feature matrix $L(g, h)$}
        Initializing $L(g, h)$ to a zero matrix

        representativeness=0
        
        \CommentSty{\%Turning angle filtering}
        
 		\If{$\Delta\theta=|\theta_{c}-\theta_{p}|>Th_{\theta}$}
		   {

                       Select the current neuron as $N_{new}$\\
              }

                           \CommentSty{\%Filtering of Feature Neurons}

		\If{representativeness $\neq 1$}
		{
  
        \CommentSty{\%Euclidean Distance Channel}
         
			\If{ $||N_{new} - N_F|| >Th_{1}$}
			{
				add $N_{new}$ into $N_{F}$\\
	          }
            \CommentSty{\%Neural Activity Channel}
            
           \For{all $N_{i}$ in $N_{F}$}
	   {

                \If{ $ \zeta_{N_{i}} < 0$}
			{
				remove $N_{i}$ from $N_{F}$\\
	          }
             $N_{D}=CheckNeighbor(N_i, N_F)$
           }

                                    \CommentSty{\%Secondary Distance Fusion}

                                    \For{all $N_{D}$}
	   {
                \If{ $||N_D -N_{D+1}||<Th_2$}
			{
				remove $N_{D+1}$ from $N_{F}$\\
    }
                }

                update the value of representativeness\\
                update $L(g, h)$ with its corresponding index\\
  }
  	\Else
		{
			optimize feature set $N_{F}$
		}
	\Return $N_{F}$ and $L(g, h)$
 	
\end{algorithm}

\subsubsection{\textbf{Extraction and Filtering of Feature Neurons}}
\begin{figure}[!t]
\centering
\includegraphics[width=3.4in]{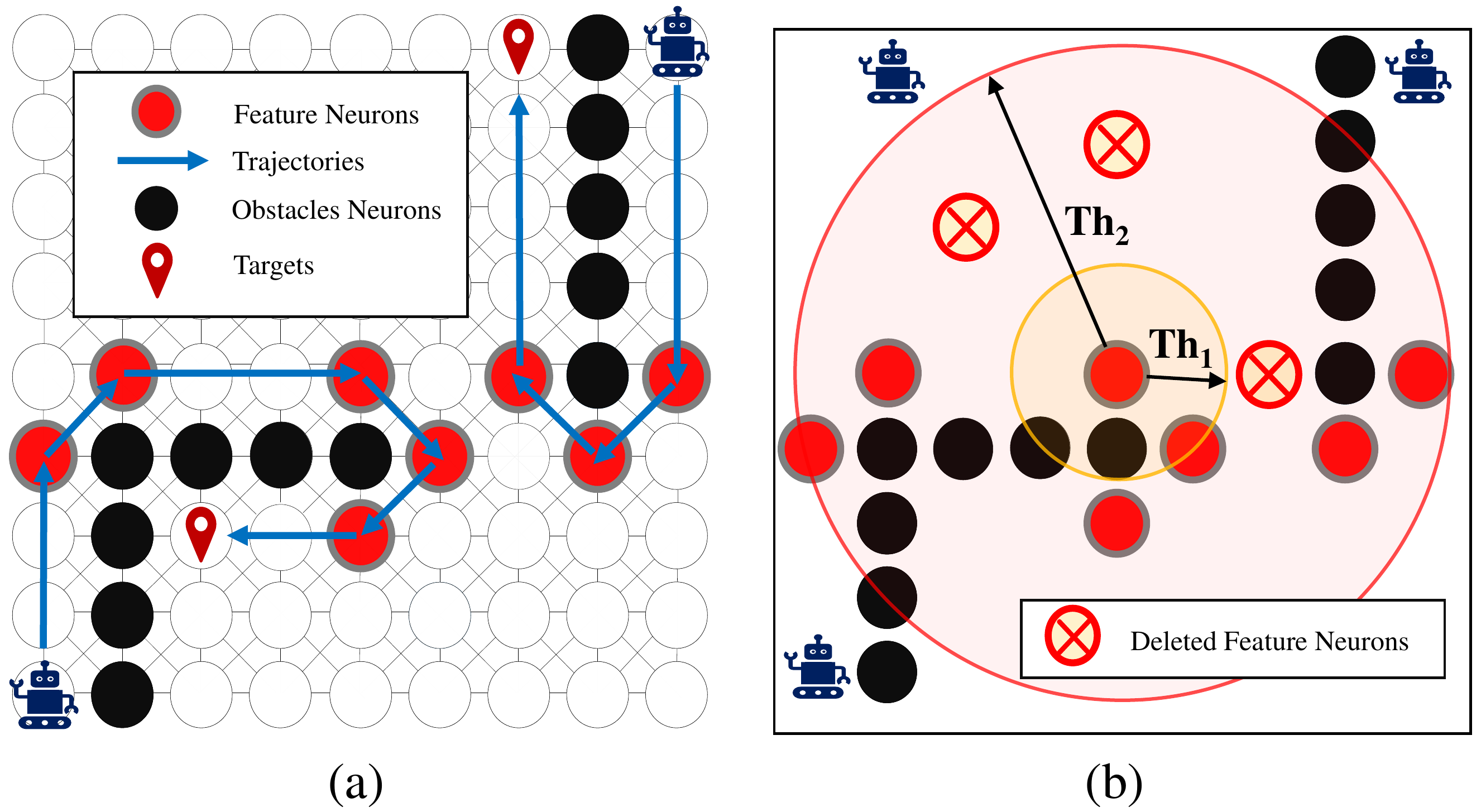}
\caption{Feature neuron extraction and filtering. (a)a schematic representation of the feature neuron extraction process; (b)the secondary distance fusion.}
\label{fig_robot_path}
\end{figure}
In this section, a dual-channel scale filter is employed to select feature neurons $N_{new}$. As shown from line 4 to line 5 in Algorithm \ref{alg:NN}, the initial channel constitutes the turning angle channel.  The turning angle filtering function can be defined as
\begin{equation}
N_{new}=\left\{(x,y), \Delta\theta >Th_{\theta}\right\},
\label{angle_filtering}
\end{equation}
where $N_{new}$ is the feature neurons selected by the turning angle channel; $Th_{\theta}$ is the angular threshold and $\Delta\theta=|\theta_{c}-\theta_{p}|$ refers to the angular deviation between the current moving direction and the preceding movement direction, respectively. If the turning angle at the current neutron is greater than a manually set angular threshold, this neuron is selected as $N_{new}$. Therefore, the feature neurons signify that the robot requires a big angular change at the current neuron. A small turning angle suggests that the robot is on a linear path. Neurons exhibiting small angular deviation can be seamlessly connected to other feature neurons.
As shown in Fig. \ref{fig_robot_path}(a), rescue paths are illustrated by the blue line, while the extracted feature neurons are denoted by red circles, and the obstacle neurons are represented by black circles.

Furthermore, an additional Euclidean distance channel is devised by incorporating the threshold of distance. The  Euclidean distance filtering function can be defined as
\begin{equation}
N_F=\left\{N_{new}, ||N_{new} - N_F|| >Th_{1}\right\},
\end{equation}
 where $N_{F}$ is the feature neurons that pass the  Euclidean distance channel and $Th_{1}$ is the pre-designed distance threshold. 
When selecting a new feature neuron $N_{new}$ from the turning angle channel, the Euclidean distance is calculated between $N_{new}$ and all other existing feature neurons $N_F$, as shown from line 8 to line 10 in Algorithm \ref{alg:NN}. If this distance exceeds the threshold $Th_{1}$, the neuron $N_{new}$ is added into the feature neurons $N_F$.


Moreover, to keep rescue safe and adapt to dynamic changes in the environment, a neural activity channel is proposed based on the dynamic landscape of the neural activity, as shown from line 12 to line 14 in Algorithm \ref{alg:NN}. The neural activity filtering function can be defined as
\begin{equation}
N_{F}=\left\{N_{F}-N_{i} , \zeta_{N_{i}} < 0\right\},
\end{equation}
 where  $\zeta_{N_{i}}$ is the neural activity of $N_{i}$ feature neurons. After passing the neural activity channel, only the feature neurons with non-negative neural activity will be reserved. Because the $\sum^n_{l=1}v_{kl}[\zeta_l-\sigma]^-$ term in (\ref{eq:binnE_safe}) allows negative activity to stay locally only in a small region, feature neurons that pass through the neural activity channel will keep a safe distance from obstacles. Furthermore, when obstacles move to a new position, neural activity becomes positive. Therefore, when the obstacle moves, the original neurons are able to participate in the selection of feature neurons.

To further minimize the number of redundant feature neurons, a secondary distance fusion approach is used to improve and streamline the set of feature neurons, as shown from line 16 to line 19 in Algorithm \ref{alg:NN}. The secondary distance fusion can be defined as 
\begin{equation}
\begin{aligned}
 N_{F} = \left\{N_D | N_D \in N_F, ||N_D -N_{D+1}||<Th_2\right\},
\end{aligned}
	\label{ratioadjust}
\end{equation}
where $Th_2$ is the secondary fusion distance threshold and $N_D$ is the feature neurons that participate in the secondary distance fusion. As shown in line 15 in Algorithm \ref{alg:NN}, each feature neuron is performed into the collision-checking function. If the collision check of neurons is passed, these two neurons are considered connected. 
Feature neurons with more than three connected neurons are extracted as elements in $N_D$. If the distance between $N_D$ and $N_{D+1}$ is smaller than  $Th_2$ and they are collision-free through the neural network, $N_{D+1}$ is removed from $N_F$. The same iteration with $N_{D+1}$ will continue to work until all the feature neurons in $N_F$ are tested, as illustrated in Fig. \ref{fig_robot_path}(b). 



\subsubsection{\textbf{Feature Representativeness and Neurodynamics-based Feature Matrix}}
Following the extraction of feature neurons, a clustering procedure is introduced to document the associations between feature neurons and their corresponding neurons. 
The representativeness value is proposed to evaluate the feature extraction of the neural network. The  representativeness value $R_{rep}$ can be defined as
\begin{equation}
R_{rep }=\frac{\operatorname{count}(\text { represented neurons })}{\operatorname{count}(\text { total neurons})},
\end{equation}
where $R_{rep}$ is defined as the ratio of represented neurons to total neurons. The neuron that has non-negative neural activity will find the nearest feature neuron that can pass the collision check.  If the nearest feature neuron can pass the collision check, the neuron with non-negative neural activity is represented by this feature neuron. 
Typically, each feature neuron responds to multiple neighboring neurons. Initially, the representative value is set to zero $(R_{rep} = 0)$, increasing throughout the feature learning process. When reaching one $(R_{rep} = 1)$, all neurons within the environment can be represented by at least one feature neuron. 


Inspired by the feature matrix of Chi\textit{ et al.} \cite{chi2021generalized}, a neurodynamics-based feature matrix is proposed to represent the original neural network. As shown in line 21 in Algorithm \ref{alg:NN}, the topological connection of the neurodynamics-based feature matrix can be obtained as
\begin{equation}
L(g,h) =\left\{ 
\begin{aligned}
 &||N_{F}(g)-N_{F}(h)||, \quad \text{if it is collision-free},   \\
&\qquad \qquad 0,  \qquad \qquad \qquad \, \,\, \, \text{otherwise},   \\
\end{aligned}
\right.
	\label{neuronconnection}
\end{equation}
where  $L(g, h)$ signifies the topological connection between the $g$-th and $h$-th feature neurons. If these neurons are connectable without collisions through neural activity, $L(g, h)$ records the distance between the feature neurons. Conversely, if the $g$-th and $h$-th feature neurons cannot be connected, the distance remains zero.
It is improtant to highlight the difference between the feature matrix proposed by Chi \textit{et al.} \cite{chi2021generalized} and the neurodynamics-based feature matrix. The collision check in the proposed neurodynamics-based feature matrix is based on the dynamic landscape of neural activity.  Therefore, by choosing an appropriate $\sigma$, the topological relationships of each feature neuron can ensure that the generated path does not cross the narrow spaces between obstacles and intersect with the corners of obstacles.  As shown in Fig. \ref{fig_path}, neurons $N_1$ and $N_2$ should be connectable based on Chi \textit{et al.} collision checking. However, considering the potential interference of obstacles, a connection between these two neurons was not established. Therefore, the generated path is longer, but keeps the robot safe.

\subsubsection{\textbf{Feature Neuron Optimization}}
The sub-optimal paths can achieve the rescue targets when all neurons within the feasible environment can be represented by at least one feature neuron $(R_{rep}=1)$. To improve the efficiency of feature neurons and rescue paths in complex environments, the feature learning process will continue when $R_{rep}=1$. As shown in line 23 in Algorithm \ref{alg:NN},  feature neuron optimization is implemented to continually improve the existing feature neurons. 
Firstly, the distance of the new featured neuron $N_{new}$ and the feature neuron in $N_{F}$  is obtained as
\begin{equation}
\label{eq:distance threshold}
\begin{aligned}
N_{I_{min}} \Leftarrow I_{min} =\arg\min_{e=1}^{K}||N_{new}-N_{F}(e)||, \\
\end{aligned}
\end{equation}
where $K$ is the number of the feature neurons; $I_{min}$ is the minimum distance to the new featured neuron $N_{new}$ and $N_{I_{min}}$ is the feature neuron with the minimum distance to the new featured neuron $N_{new}$. All neurons that are represented by the feature neuron $N_{I_{min}}$ are denoted as $\mathcal{N}_{I_{min}}$.
The evaluation process involves testing whether the new feature neuron $N_{new}$ passes collision detection with neurons in $\mathcal{N}_{I_{min}}$. Since the connection relationship has been recorded in the neurodynamics-based feature matrix, the collision check process could be efficient during optimization. If no collision occurs, the average distance between the feature neuron and its represented neurons is calculated to determine whether to replace the feature neuron $N_{I_{min}}$, which can be defined as
\begin{equation}
\label{eq:average_distance}
\begin{aligned}
d(N_k)=\frac{1}{\operatorname{count}(\mathcal{N}_{I_{min}})}\sum_{n_i\in\mathcal{N}_{I_{min}}}||N_k-N_i||,\\
\end{aligned}
\end{equation}
where $\operatorname{count}(\mathcal{N}_{I_{min}})$ denotes the total number of neurons in $\mathcal{N}_{I_{min}}$; $N_k$ is $k$-th feature neuron and $N_i$ is $i$-th represented neuron in $\mathcal{N}_{I_{min}}$. If $d(N_{new})<d(N_{I_{min}})$, the feature neuron $N_{I_{min}}$ will be replaced with $N_{new}$. This optimization procedure remains iterative throughout the process.  Note that this comparison considers only the represented neurons $\mathcal{N}_{I_{min}}$, rather than the entire neural network. In practical tests, due to the cooperation of multiple robots, each robot shares the value of $d(N_{I_{min}})$  with other robots during updates or calculations, which avoids redundant computations and reduces computational load. Therefore, with the increase of robots successfully rescuing targets, there is a concomitant improvement in the efficiency and representativeness of feature neurons.

\subsection{Fast Path Planning via Neurodynamics-based Feature Matrix}
\begin{figure}[!t]
\centering
\includegraphics[width=1.8in]{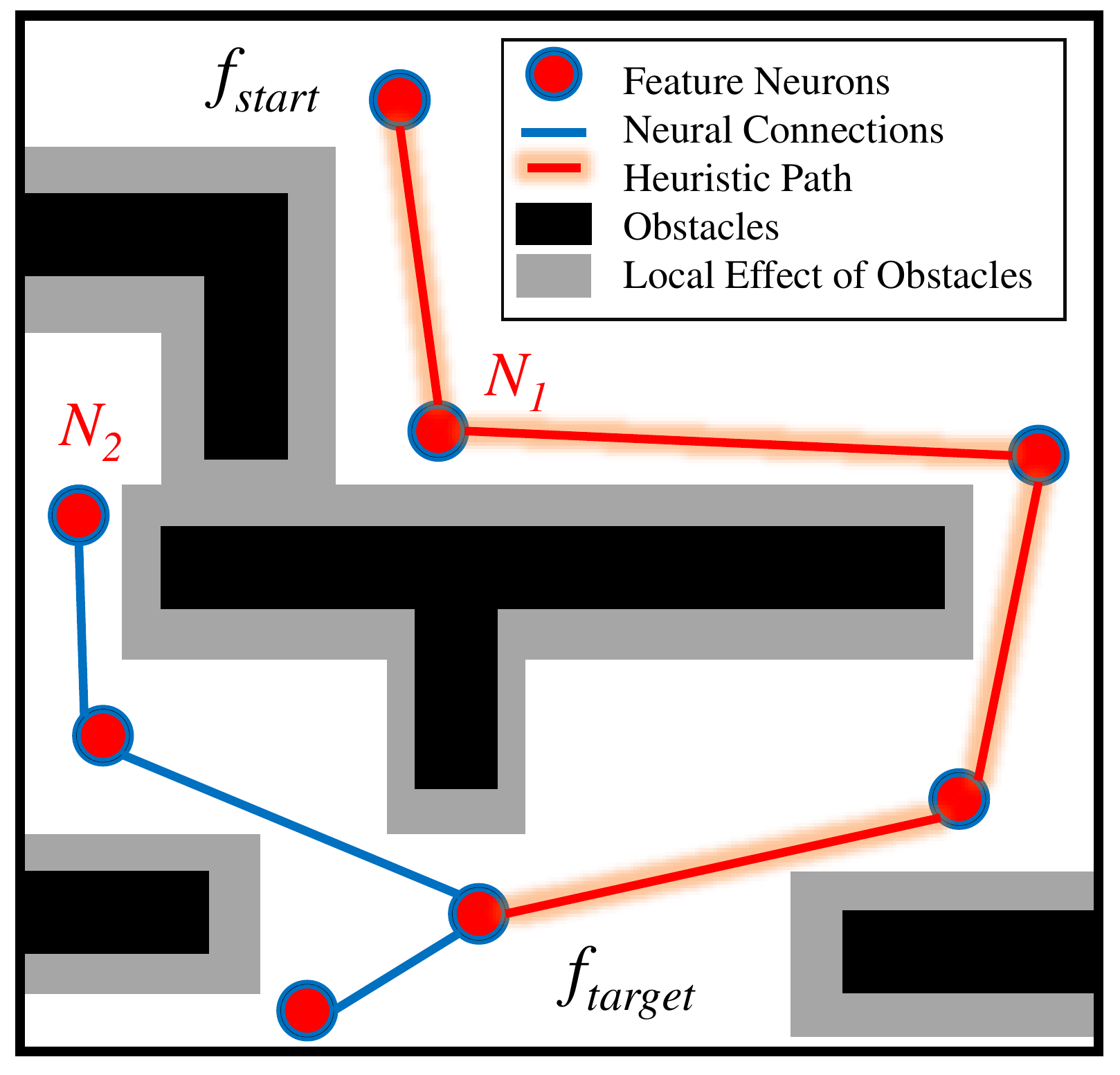}
\caption{An illustration of the heuristic path planning with parameter-driven topological adaptability.}
\label{fig_path}
\end{figure}

\begin{algorithm}[t]
\footnotesize
	\LinesNumbered
	\caption{Path Planning via Neurodynamics-based Feature Matrix}
	\label{alg:matrix}
	
	\KwIn{Start position $f_{start}$, target position $f_{target}$ and neurodynamics-based feature matrix $L(g, h)$}
	\KwOut{Collsion-free path with  topological adaptability$P$}
 
  Initializing $C(f_{start})=0$
  
     $C(f_{neig})=+\infty$
     
    $P.init=f_{target}$
    
     $f_{curr}=f_{target}$
     
    \While {$f_{curr}\neq f_{start}$}
    {
         
         $N_{neig}=find(L(f_{curr}, f_{neig})>0)$

         \For{all $f_{neig}$ in $N_{neig}$}
	{
              $C_{dist}= C(f_{curr})+L(f_{curr},f_{neig})$
              
              \If{$C_{dist}< C(f_{neig})$}
		   {
   	       $C(f_{neig})=C_{dist}$
           
                  $f_{neig}.parent \rightarrow f_{curr}$

                  $f_{curr}=f_{neig}$
                  
                  update $P$
                       
              }
	}
    }
        
\Return $P$

\end{algorithm}

Upon completion of the feature learning process $(R_{rep}=1)$, a neurodynamics-based feature matrix is established to represent the initial BINN. Using the neurodynamics-based feature matrix, a heuristic path can be generated. As shown in line 4 in Algorithm \ref{alg:matrix},  the neighboring neurons of $f_{curr}$ can be extracted from the neurodynamics-based feature matrix by searching for the value $L(f_{curr}, f_{neig})>0$, which is denoted as $N_{neig}$.  For each neighboring neuron $f_{neig} \in N_{neig}$  the cumulative cost, denoted as $C_{dist}$, is computed as follows
	\begin{equation}
	\label{eq:cost}
	\begin{aligned}
	C_{dist}= C(f_{curr})+L(f_{curr},f_{neig}).\\
	\end{aligned}
	\end{equation}
As shown from line 5 to line 11 in Algorithm \ref{alg:matrix}, if $C_{dist}$ is less than the original $C(f_{neig})$, the value of $C(f_{neig})$ is replaced by $C_{dist}$ and the parent node of $f_{neig}$ is set as $f_{curr}$. The path $P$ is updated with a new connection from $f_{curr}$ to $f_{neig}$. A visual representation of the heuristic path planning can be observed in Fig. \ref{fig_path}. Given a specified target pose, the corresponding feature neuron $f_{target}$ is obtained from the neurodynamics-based feature matrix. A heuristic path can be directly produced by performing a backward search from $f_{target}$ to $f_{start}$, as demonstrated by the red path in Fig. \ref{fig_path}.
 From Algorithm \ref{alg:matrix}, the most frequent statement in this algorithm is in line 6, therefore, the complexity of the program is $O(n^2)$ since the connection relationship has been recorded in the neurodynamics-based feature matrix, no additional collision checks between the feature neurons are needed, which reduces the algorithm complexity to $O(n^2)$ \cite{chi2021generalized}. 

\section{Simulation Results}
\label{sec:simulation}

The simulation test setting involves using robots to rescue multiple targets in four dynamic and complex environments: static obstacles, moving obstacles, sudden obstacles, and house-like scenarios. In all simulation studies, MATLAB R2021a serves as the testing platform. The parameters for these simulations are established as follows: $A=5$, $B=1$, $D=1$, $\mu=1$, $E=70$, $\sigma=-0.5$, $L_n=1m$, $r_0=\sqrt{2}$, $Th_1=3$, and $Th_2=5$.  The size of BINN has $70 \times 70$ neurons. In order to evaluate the efficacy and efficiency of the proposed methodology, it is compared to the conventional BINN.

\subsection{Rescue Performance in a Static Environment}

\begin{figure}[htbp]
\centering
\includegraphics[width=3in]{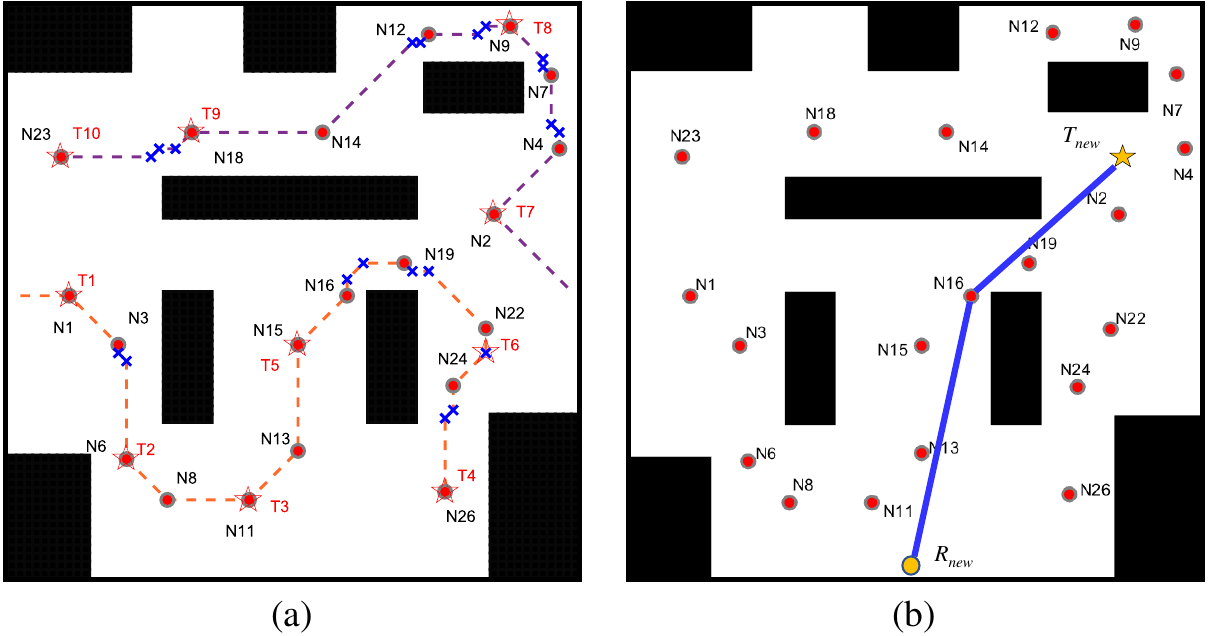}
\caption{The rescue performance in a static environment. (a) feature neurons; (b) the heuristic path.}
\label{static_show}
\end{figure}

The proposed methodology is initially implemented for a rescue task in a static obstacle environment without traps. Fig.\ref{static_show}(a) illustrates the trajectories generated by BINN along with the extracted feature neurons. The initial positions of the robots are (1,35) and (70,35), while the target locations are (8,35), (15,15), (30,10), (54,11), (36,29), (59,28), (60,45), (63,67), (23,55) and (7,52). Fig. \ref{static_show}(b) presents an exemplary heuristic path for the new robot, denoted as $R_{new}$, and a new target, denoted as $T_{new}$. The positions of the new robot and target are (35,2) and (60,52), respectively. It is evident that the heuristic path represents the shortest connection from the starting point to the feature neuron $N_{16}$ and the target while ensuring obstacle avoidance.

\subsection{Rescue Performance with Moving Obstacles}

\begin{figure}[htbp]
\centering
\includegraphics[width=3in]{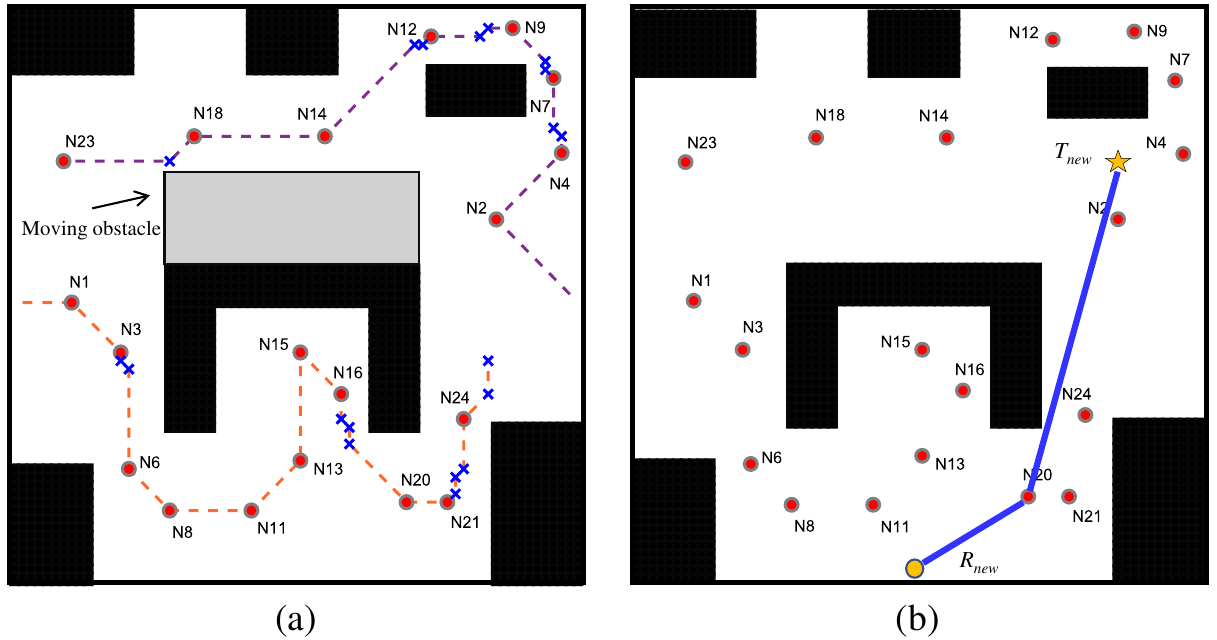}
\caption{The rescue performance with moving obstacles. (a) feature neurons; (b) the heuristic path.}
\label{dynamic_show2}
\end{figure}

The next simulation is applied to a more complex scenario, where one obstacle is moving. 
The initial locations of the robots and the target remain consistent with the first simulation. Moving obstacles move downward and stop at positions ranging from (20,45) to (50,39). 
As shown in Fig. \ref{dynamic_show2}(a), the moving obstacles fully block the gap before the robot navigates through it and thereby form U-shaped obstacles. 
The robot has to move away from the target and pass through the U-shaped obstacles. The generated trajectory effectively avoids obstacles extending from the initial position to the target.  As shown in Fig. \ref{dynamic_show2}(b), the generated heuristic path connects the feature neuron $N_{20}$ and is able to pass around the U-shaped obstacles. 
\subsection{Rescue Performance with Sudden-Obstacle Changes}
\begin{figure}[htbp]
\centering
\includegraphics[width=3in]{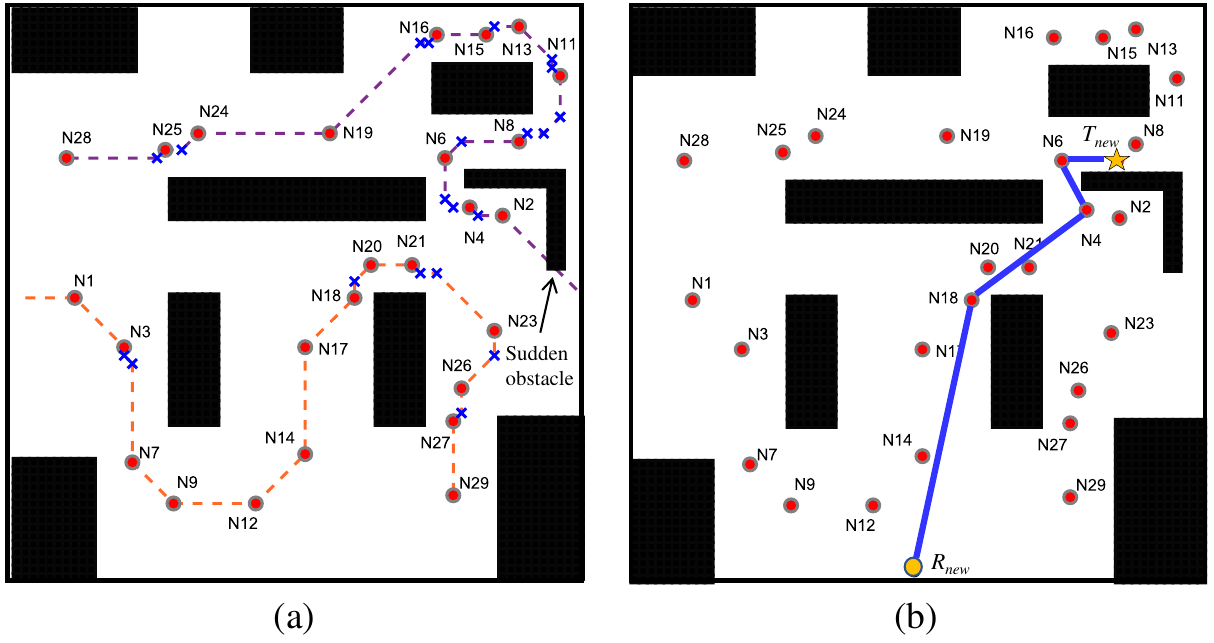}
\caption{The rescue performance with sudden-obstacle changes. (a) feature neurons; (b) the heuristic path.}
\label{sudden_show}
\end{figure}

The next simulation is applied to a more dynamic scenario with some sudden environmental changes. The initial positions of the robots and target are the same as in the first simulation. 
In addition, when the robot moves toward the target, the L-shaped obstacles are suddenly placed in front of the robot, as shown in Fig. \ref{sudden_show}(a). It shows that the robot begins to leave away from the target when the L-shaped obstacles are suddenly placed, then passes around these obstacles. The heuristic path is shown in Fig. \ref{sudden_show}(b), which ultimately reaches the target while avoiding any collisions.
\subsection{Rescue Performance in a House-Like Environment}
\begin{figure}[htbp]
\centering
\includegraphics[width=3in]{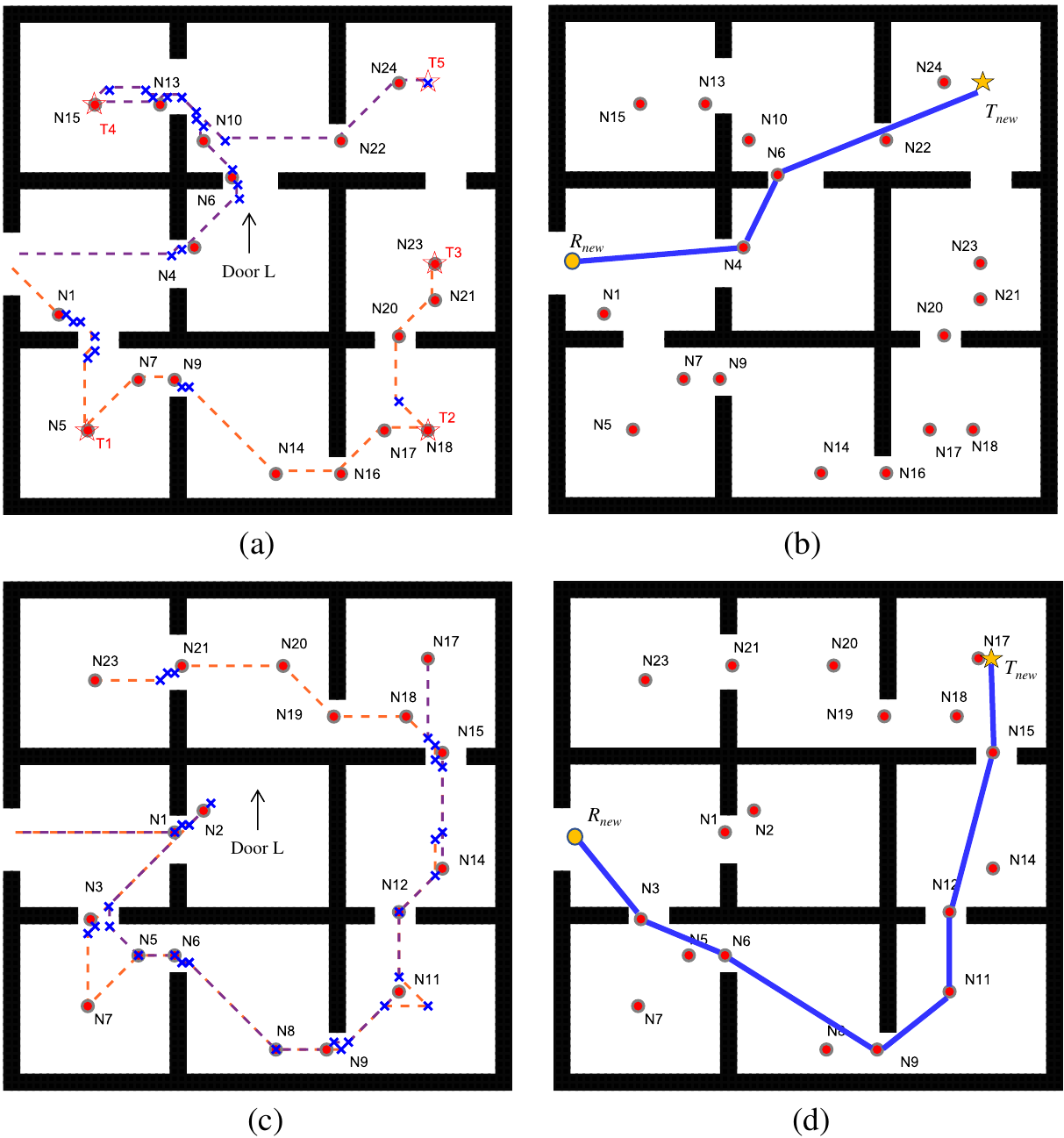}
\caption{The rescue performance in a house-like environment. (a) feature neurons when Door L is opened; (b) the heuristic path when Door L is opened; (c) feature neurons when Door L is closed; (d) the heuristic path when Door L is closed.}
\label{house}
\end{figure}

The final simulation is conducted within a complex house-like environment characterized by multiple deadlock situations, which may potentially trap the robot.  The target locations are (12,12), (59,12), (60,35), (13,57), and (59,60). Fig. \ref{house} presents this environment, where doors possess the ability to be opened or closed. In the scenario where Door L is open, the generated trajectory and heuristic path are shown in Fig. \ref{house}(a)-(b); the robot moves toward the target while maintaining a secure distance from the obstacles. On the contrary, when door L is closed, the generated trajectory and heuristic path are shown in Fig. \ref{house}(c)-(d). In this situation, the robot must traverse a considerably longer route to reach the target. It is important to note that the rescue process does not have learning procedures.

\subsection{Comparison Studies of Rescue Performance}
In comparison studies, the proposed FLBBINN is compared with the traditional BINN in various scenarios. 
In these tests, the positions of robots, targets, and obstacles are randomly distributed in the environment. 
It is evident that the proposed FLBBINN consistently outperforms the traditional BINN method in all scenarios, as shown in Table \ref{table_compare}. The proposed FLBBINN uses significantly fewer neurons compared to BINN, ranging from $18$ to $26$ neurons as opposed to $4900$ neurons in the BINN method. Additionally, the proposed FLBBINN generated shorter paths in all scenarios. The planned path based on traditional BINN is the shortest connection between neurons under grid discretization, whereas the path generated by FLBBINN is based on the actual shortest path between the feature neuron and the robot. Finally, the proposed FLBBINN is capable of dealing with the slow propagation problem. It is important to note that there is a time cost of $110.3s$ as opposed to $143.3s$ in the house-like scenario when the door is closed. Due to the presence of only local connections, the propagation time increases significantly when obstacles intervene between the two positions. Especially within enclosed complex environments, the robot remains in a prolonged idle state. When the door is closed, the neural activity needs to travel a very long distance before it can propagate from the target point to the robot. As a result, the robot remains in a prolonged idle state. In the comparison, the robot based on traditional BINN waits for $10.1s$ to stay at the start position.

\begin{table}[htbp]   
\centering
\caption{The performance comparison of the FLBBINN and traditional BINN} 
\scalebox{0.9}{
\begin{tabular}{ccccc}    
\toprule
 Scenarios& Methods & Neurons & Path Length(m) & Time(s)\\    
\midrule   
\multirow{2}{*}{Static}& BINN & $4900$ & $60.4$ & $52.6$\\  
&FLBBINN & $ 18$  &  $ 49.5$ & $35.4$\\  
\midrule  
\multirow{2}{*}{Moving}& BINN & $ 4900$ & $60.4$ & $51.9$\\  
&FLBBINN & $ 19$  &  $ 59.1$& $42.2$\\ 
\midrule 
\multirow{2}{*}{Sudden}& BINN & $4900$ & $65.8$ & $57.4$\\ 
&FLBBINN & $ 26$  &  $ 65.2$& $46.6$\\ 
\midrule  
House-like& BINN & $ 4900$ & $69.4$ & $60.3$\\ 
(door open)&FLBBINN & $ 18$  &  $ 67.5$& $48.2$\\ 
\midrule  
House-like& BINN & $ 4900$ & $149.3$ & $ 143.3$\\ 
(door close)&FLBBINN & $ 18$  &  $ 148.6$& $ 110.3$\\ 
\bottomrule   
\label{table_compare}
\end{tabular} 
}
\end{table}

\section{Real-Robot Experiments}
\label{sec:experiments}
\begin{figure}[!t]
\centering
\includegraphics[width=2.5in]{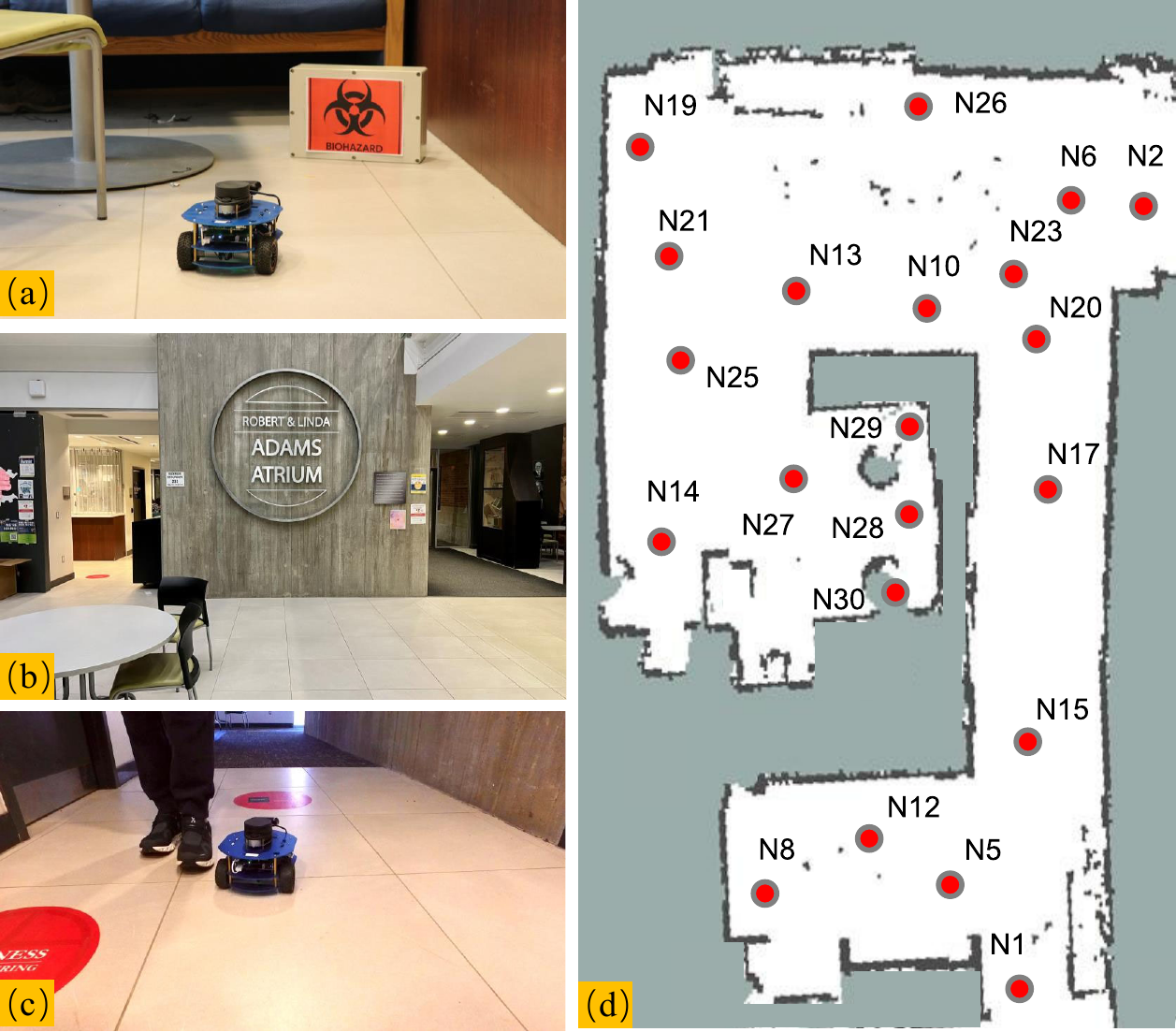}
\caption{The illustration of the real-world experiment. (a) the ROS-based mobile robot; (b) the experiment is within an area of $35m$ $\times$ $20m$; (c) the mobile robot avoids collision with the pedestrian; (d) the robot-generated 2D occupancy grid map. The red points indicate the feature neurons. }
\label{fig_experiment}
\vspace{-0.5cm}
\end{figure}

The  proposed  rescue framework was tested using multiple mobile robots in a real-world environment. Three Robot Operating System (ROS) robots were built for experimental purposes. As shown in Fig. \ref{fig_experiment}(a), 
robot dynamics is described by means of differential drive models. Each robot carried a 1080p camera implemented in Ubuntu using the OpenCV library and one RPLIDAR A1 laser scanner, which has $12m $ and $360\degree$ omnidirectional range scanning. 
As shown in Fig. \ref{fig_experiment}(b), the experiment is within an area of $35m$ $\times$ $20m$ where the environment is designed to include a multitude of obstacles and narrow spaces. The test environment incorporates dynamic elements, such as moving objects and people, as shown in Fig. \ref{fig_experiment}(c). The target is chosen as a box with red biohazard labeling. Fig. \ref{fig_experiment}(d) shows the experimental result of the proposed FLBBINN.

\subsection{Efficiency of the Proposed Approach}
Table \ref{table_real} shows the performance comparison of the real-world experiment. The results show that the proposed FLBBINN significantly outperforms the traditional BINN method in all performance metrics. Firstly, the experimental area is scanned as $103$ $\times$ $324$ nodes. In traditional BINN, the environment is represented one-to-one with the neural network. Therefore, the total number of neurons is $33,372$. The proposed FLBBINN uses only $21$ feature neurons, while the BINN method requires a considerably larger number of neurons, totaling $33,372$. Furthermore, the proposed FLBBINN also demonstrates better optimality by providing a shorter path length of $45.5m$, compared to $49.2m$ achieved by the BINN method.  Finally, the time taken for the proposed FLBBINN is considerably lower, taking only $25.2s$, whereas the BINN method requires $75.4s$. This is because the real-world environment is very complex and the size of the neural work is large. Therefore, it costs $45.3s$ for the robot to remain in the starting position to wait for the propagation of neural activity.
\begin{table}[htbp]   
\centering
\caption{The performance comparison of the real-world experiment } 
{
\begin{tabular}{cccc}    
\toprule
 Methods & Neurons & Path Length(m) & Time(s)\\    
\midrule   
 BINN & $33,372$ & $49.2$ & $75.4$\\  
FLBBINN & $21$  &  $45.5$& $25.2$\\  
\bottomrule   
\label{table_real}
\end{tabular} 
}
\vspace{-0.7cm}
\end{table}


\subsection{Discussions of Environmental Feature Learning}
The comparison with other feature learning methods proposed by Yuan \textit{et al.} \cite{yuan2022gaussian} is discussed to verify the effectiveness of the proposed approach. 
First, the FLBBINN feature learning process is performed automatically, while Yuan \textit{ et al.} method depends on manually setting targets and controlling the robot movement to learn about the environment. 
As shown in Table \ref{table_FL}, 
Yuan \textit{et al.}'s method has a success rate of $13/20$, because its method is hard to adapt to dynamic and sudden environmental changes. Meanwhile, the robot might have collisions with obstacles in a narrow space.
 There is no clearance from the obstacles, when choosing $ \sigma= -1.5$. Therefore, the path length is shorter than others, but the robot might have collisions with obstacles.  When choosing $\sigma = -0.1$, it shows a very strong clearance from obstacles, which pushes the robot very far from the obstacle and increases the path length. Although in the experiments various $\sigma$ ensured the completion of the rescue task, selecting the appropriate parameter $\sigma$ remains crucial in complex real-world environments. Therefore, FLBBINN ($\sigma=-0.5$) shows a compelling balance between path length and safe consideration.

\begin{table}[htbp]   
\centering
\caption{The performance comparison of Yuan \textit{et al.} method and FLBBINN }
{
\begin{tabular}{ccc}    
\toprule
 Methods  & Travel Distance(m) & Success Rate\\    
\midrule   
 Yuan \textit{et al.}\cite{yuan2022gaussian}   & $42.3$ & $13/20$\\  
FLBBINN ($\sigma=-1.5$)  &  $41.6$& $20/20$\\
FLBBINN ($\sigma=-0.5$)  &  $45.5$& $20/20$\\ 
FLBBINN ($\sigma=-0.1$)  &  $58.2$& $20/20$\\ 
\bottomrule   
\label{table_FL}
\end{tabular} 
}
\vspace{-0.5cm}
\end{table}


\subsection{Parameter Sensitivity}
\begin{figure}[!t]
\centering
\includegraphics[width=2.3in]{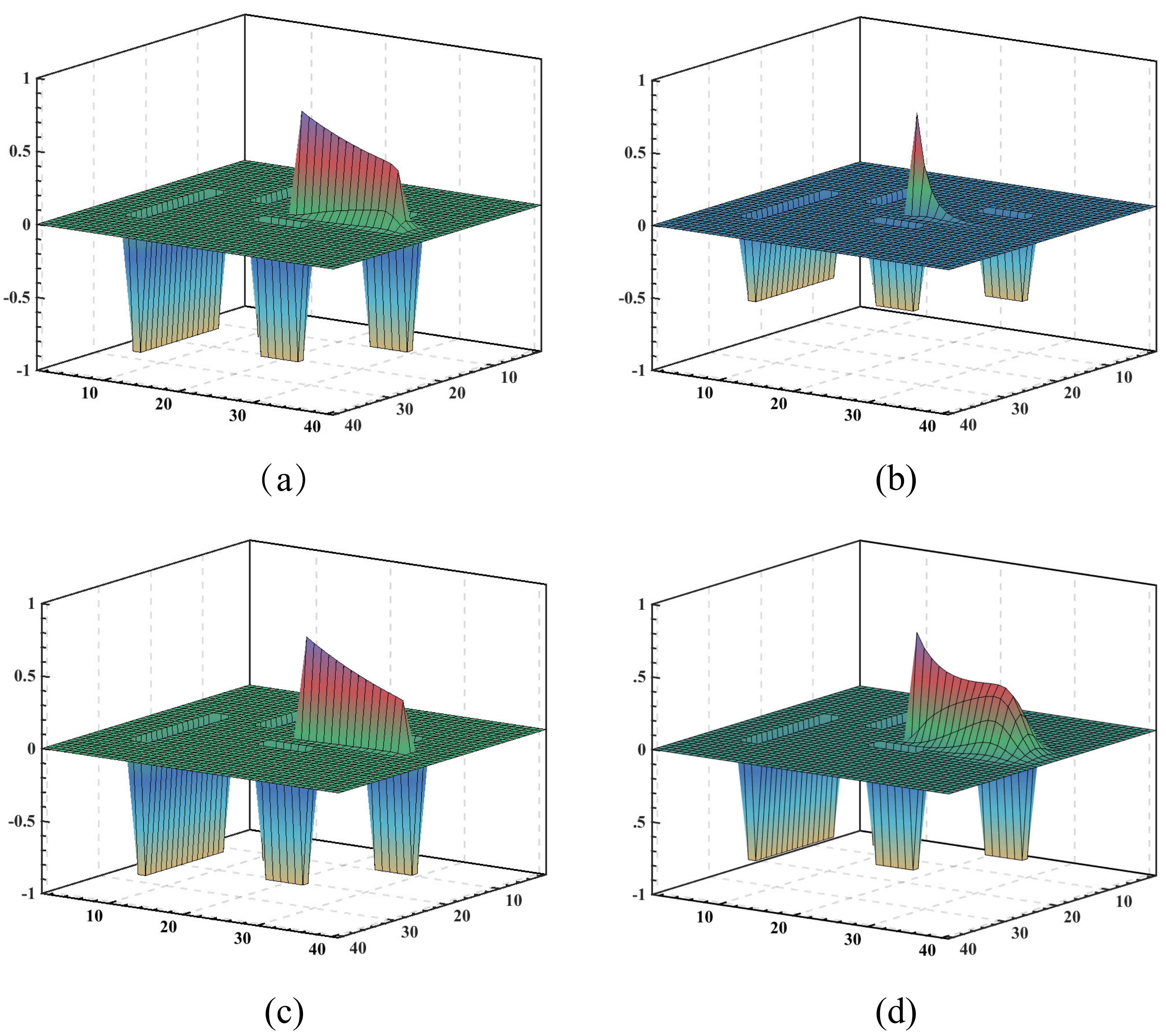}
\caption{The neural activity when choosing different $A$ and μ values. (a) $A=10$ and $\mu=1$; (b) $A=30$ and $\mu=1$; (c) $A=10$ and $\mu=0.1$; (d) $A=10$and $\mu=5$. }
\label{para}
\vspace{-0.5cm}
\end{figure}

The proposed approach aimed to provide real-world rescue applications. Therefore, most parameters are decided by real-world applications, such as the robot velocity and thresholds for the extraction of feature neurons. The parameters of the shunting equation have been discussed in previous work \cite{Li2021bio}. In this discussion, the most important parameters $A$ and $\mu$ are discussed, and some simulations are carried out to demonstrate the effect of these parameters. To analyze the influence of parameter $A$, several simulations are tested with the same parameter settings, except that $A$ has different values. Figs. \ref{para}(a) and \ref{para}(b) show two landscapes of neural activity when choosing $A =10$ and $A =30$. The small transient response of activity makes the past influence of external inputs disappear slowly. When choosing a bigger $A$ value, the activity propagation from target becomes the domain contribution to the formation of the neuron activities. To analyze the influence of the parameter $\mu$, several experiments are tested with the same parameter settings, except that $\mu$ has a different value. The propagation of the neural activity from the target is weakened because of choosing a smaller $\mu$ value in Fig. \ref{para}(c). The remaining neural activity has a relatively stronger influence on the neural network. In addition, when parameter $\mu$ > 1, the neural network is very easy to saturate because the propagated activity is amplified, as shown in Fig. \ref{para}(d). Thus, the value of $\mu$ is normally selected in the interval $\mu \in (0,1]$.
\section{Conclusion}
\label{sec:conclusion}
In this paper, a novel FLBBINN is proposed for executing rescue operations in complex and dynamic environments. Upon the culmination of the feature learning, a neurodynamics-based feature matrix is constituted. The proposed neurodynamics-based feature matrix facilitates the rapid generation of heuristic rescue trajectories with parameter-driven topological adaptability. Extensive simulations and real-robot experiments are conducted to evaluate and verify the proposed FLBBINN performance in terms of speed, efficiency, and optimal performance. 
Potential work could be to integrate multiple sensing modalities to improve the situational awareness of robots and enable them to navigate complex environments more effectively. In addition,  the future work will include to use distributed computing to lessen computational load or adopt innovative optimization algorithms to improve the effectiveness of feature neurons.


\bibliographystyle{IEEEtranTIE}
\bibliography{reference_yuan_test}\ 

\vspace{-2.0cm}
    \begin{IEEEbiography}[{\includegraphics[width=1in,height=1.25in,clip,keepaspectratio]{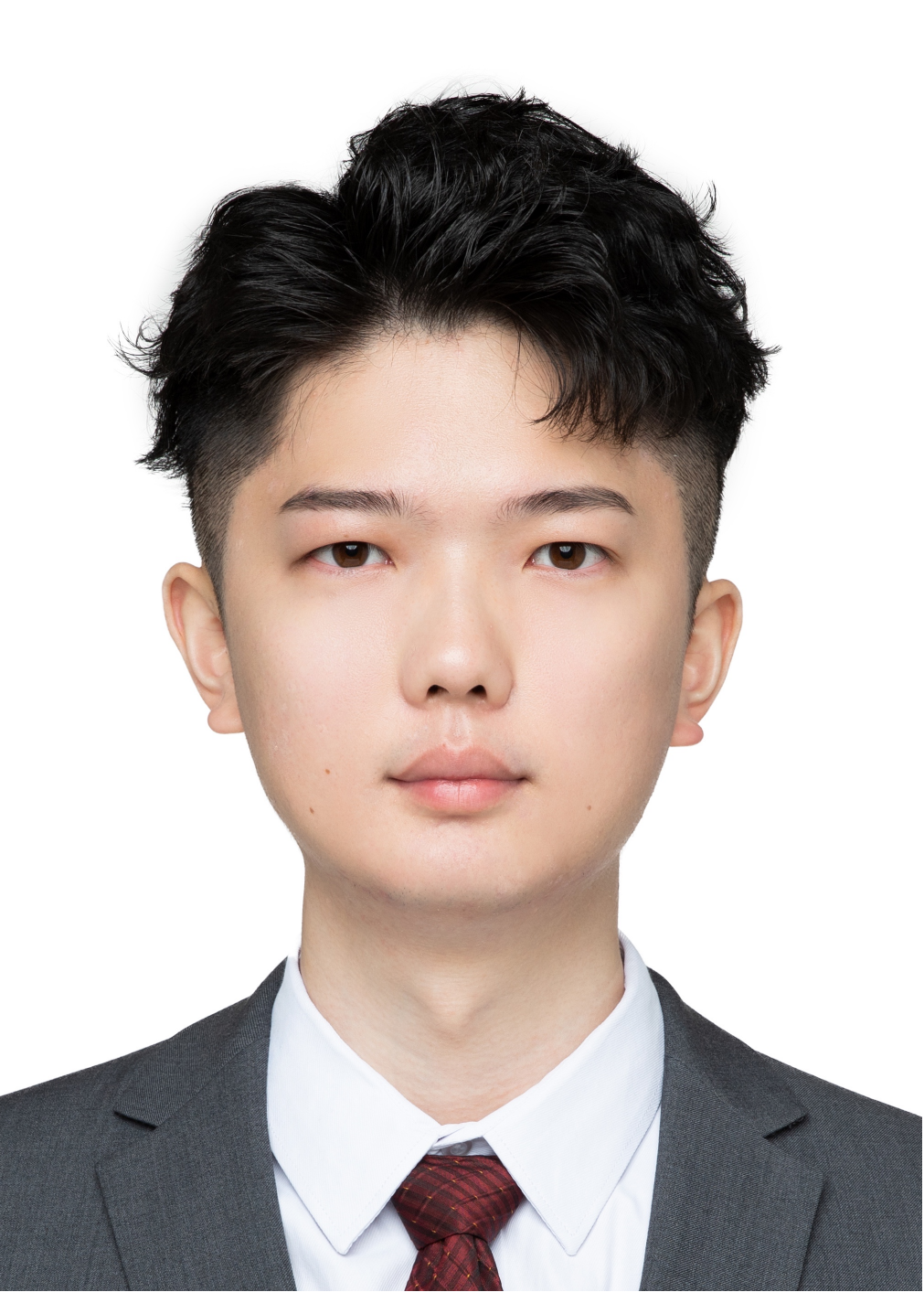}}]{Junfei Li} (Member, IEEE)  received the B.Eng. degree in communication engineering from Chongqing University of Posts and Telecommunications, Chongqing, China, in 2017, and the Ph.D. degree in engineering systems and computing from the University of Guelph, Ontario, Canada, in 2023. He is currently a Postdoctoral Research Fellow at the School of Engineering, University of Guelph, Ontario, Canada. His research interests include escape behaviors, search and rescue, and bio-inspired algorithms.


\end{IEEEbiography}

\vspace{-3.0cm}
\begin{IEEEbiography}[{\includegraphics[width=1in,height=1.25in,clip,keepaspectratio]{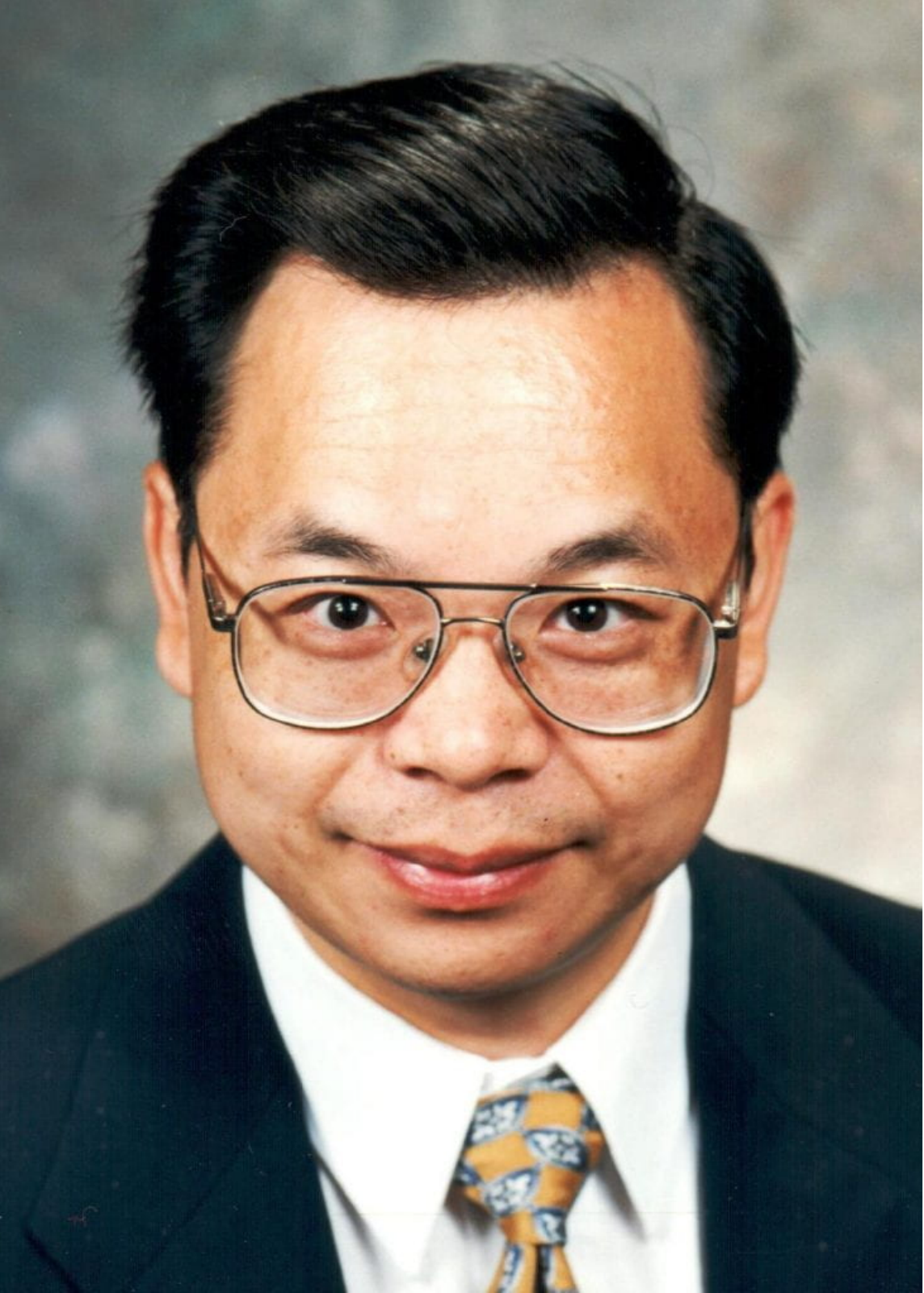}}]
{Simon X. Yang} (Senior Member, IEEE)  received the B.Sc. degree in engineering physics from the Beijing University, Beijing, China, in
 1987, the first of two M.Sc. degrees in biophysics from the Chinese Academy of Sciences, Beijing, China, in 1990, the second M.Sc. degree
 in electrical engineering from the University of Houston, Houston, TX, USA, in 1996, and the Ph.D. degree in electrical and computer engineering from the University of Alberta, Edmonton, AB, Canada, in 1999.
 
 He is currently a Professor and the Head of the Advanced Robotics
 and Intelligent Systems Laboratory with the University of Guelph,
 Guelph, ON, Canada. His research interests include robotics, intelligent systems, sensors and multisensor fusion, wireless sensor networks, control systems, machine learning, fuzzy systems, and computational neuroscience. Dr. Yang has been very active in professional activities. He serves as the Editor-in-Chief of International Journal of Robotics and Automation, and an Associate Editor of IEEE TRANSACTIONS ON CYBERNETICS, IEEE TRANSACTIONS ON ARTIFICIAL INTELLIGENCE, and several other journals. He was involved in the organization of many international conferences.
\end{IEEEbiography}




\end{document}